\documentclass[runningheads]{llncs}
\usepackage{graphicx}
\usepackage[inkscapearea=page]{svg}
\usepackage{csquotes}
\usepackage{tikz}
\usetikzlibrary{calc}
\tikzset{pics/folder/.style={code={%
    \node[inner sep=0pt, minimum size=#1](-foldericon){};
    \node[folder style, inner sep=0pt, minimum width=0.3*#1, minimum height=0.6*#1, above right, xshift=0.05*#1] at (-foldericon.west){};
    \node[folder style, inner sep=0pt, minimum size=#1] at (-foldericon.center){};}
    },
    pics/folder/.default={20pt},
    folder style/.style={draw=foldercolor!80!black,top color=foldercolor!40,bottom color=foldercolor}
}

\usepackage{multirow}

\usepackage{algorithm}
\usepackage{algpseudocode}

\usepackage{orcidlink}

\definecolor{bg}{RGB}{255, 255, 255}
\definecolor{Tumor}{RGB}{206, 54, 171}
\definecolor{Dermis}{RGB}{129, 164,  85}
\definecolor{Subcutis}{RGB}{127 ,  4 ,143}
\definecolor{Epidermis}{RGB}{71, 100, 140}
\definecolor{Inflamm-Necrosis}{RGB}{212, 108 , 44}
\definecolor{myred}{RGB}{255,0,0}
\definecolor{mygreen}{RGB}{0,255,0}
\definecolor{myblue}{RGB}{0,0,255}

\algblock{Input}{EndInput}
\algnotext{EndInput}
\algblock{Output}{EndOutput}
\algnotext{EndOutput}
\newcommand{\Desc}[2]{\Statex \hspace{1em} \makebox[4em][l]{#1}#2}

\algrenewcommand{\algorithmiccomment}[1]{{\color{gray}\hfill\texttt{\# #1}}}

\begin{document}
\title{NearbyPatchCL: Leveraging Nearby Patches for Self-Supervised Patch-Level Multi-Class Classification in Whole-Slide Images}
%
%


\author{
Gia-Bao Le\inst{1,2}\thanks{Both authors contributed equally}\orcidlink{0009-0002-2815-6798}
\and
Van-Tien Nguyen\inst{1,2}$^\star$\orcidlink{0009-0006-3983-0281}
\and
Trung-Nghia Le\inst{1,2}\thanks{Coresponding author}\orcidlink{0000-0002-7363-2610}
\and
Minh-Triet Tran\inst{1,2}\orcidlink{0000-0003-3046-3041}
}
\authorrunning{G.-B. Le et al.}
%
\institute{
University of Science, VNU-HCM, Vietnam
\and
Vietnam National University, Ho Chi Minh City, Vietnam
}
\maketitle              
\begin{abstract}
Whole-slide image (WSI) analysis plays a crucial role in cancer diagnosis and treatment. In addressing the demands of this critical task, self-supervised learning (SSL) methods have emerged as a valuable resource, leveraging their efficiency in circumventing the need for a large number of annotations, which can be both costly and time-consuming to deploy supervised methods. Nevertheless, patch-wise representation may exhibit instability in performance, primarily due to class imbalances stemming from patch selection within WSIs. In this paper, we introduce Nearby Patch Contrastive Learning (NearbyPatchCL), a novel self-supervised learning method that leverages nearby patches as positive samples and a decoupled contrastive loss for robust representation learning. Our method demonstrates a tangible enhancement in performance for downstream tasks involving patch-level multi-class classification. Additionally, we curate a new dataset derived from WSIs sourced from the Canine Cutaneous Cancer Histology, thus establishing a benchmark for the rigorous evaluation of patch-level multi-class classification methodologies. Intensive experiments show that our method significantly outperforms the supervised baseline and state-of-the-art SSL methods with top-1 classification accuracy of 87.56\%. Our method also achieves comparable results while utilizing a mere 1\% of labeled data, a stark contrast to the 100\% labeled data requirement of other approaches. Source code: \url{https://github.com/nvtien457/NearbyPatchCL}

\end{abstract}
\section{Introduction}

\begin{figure}[t!]
    \centering
    \includegraphics[width=0.7\linewidth]{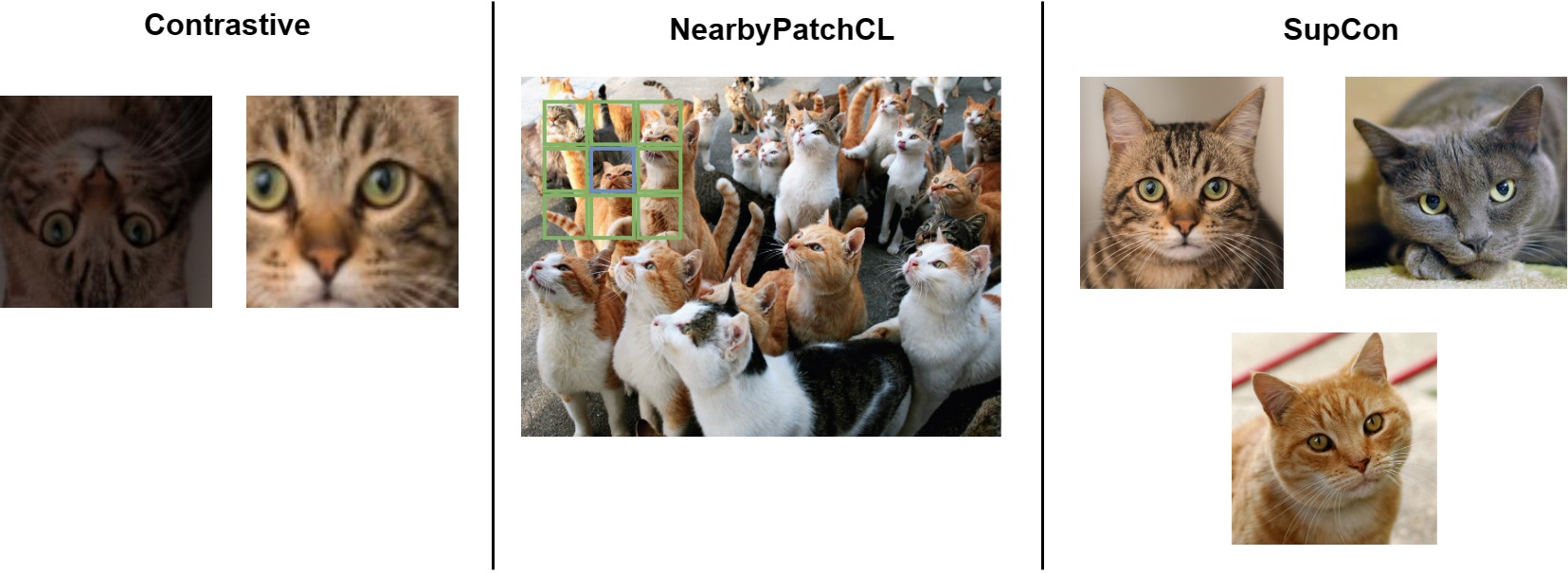}
    \caption{In contrastive learning~\cite{simclr}, our NearbyPatchCL, and SupCon \cite{SupCon}, positive samples are handled differently. Contrastive learning involves augmented pairs from the same image. In SupCon, it includes all images with the same label, while NearbyPatchCL defines positive samples that encompass all views of the center patch (blue border) and nearby patches (green border).}
    \label{fig:positive}
\end{figure}

Histology image analysis is crucial for understanding biological tissues. As Whole-Slide Images (WSI) become more common, and as cost-effective storage and fast data transfer networks become available, the creation of large databases of digitized tissue sections in hospitals and clinics is on the rise. Currently, the computational pathology community is actively digitizing tissue slides into WSI for automated analysis. There's a growing focus on developing precise algorithms for clinical use, with recent advancements in deep learning making automatic analysis of WSIs, whether supervised or weakly supervised, more popular \cite{intro:Campanella,intro:vanderLaak2021DeepLI,sup:wang2016deep,intro:Wulczyn,intro:Zheng2019EncodingHW}.

While there has been notable progress in the automated processing and clinical use of WSIs, challenges persist due to their gigapixel size. These challenges often require the use of tile-level processing and multiple instance learning for predicting clinical endpoints~\cite{intro:Wulczyn,intro:Zheng2019EncodingHW,intro:lu2021data}. Additionally, the large size of WSIs makes the annotation process by human experts cumbersome, requiring annotated data for algorithm development. The shift to deep learning further emphasizes the importance of annotations, but some methods explore the use of pre-trained representations, typically from ImageNet, as an alternative to generating WSI-specific representations \cite{intro:gamper2021multiple,li2021dual,liu2023multiple}.

Self-supervised learning (SSL) methods~\cite{simclr,azizi2021big,C3:ciga2021self} are gaining popularity for their ability to acquire competitive, versatile features compared to supervised methods. SSL involves two steps: unsupervised pre-training on unlabeled data and supervised fine-tuning on downstream tasks with limited labeled data. These methods not only require a small amount of labeled data but also enhance model performance across various histology pathology tasks \cite{C3:ciga2021self,c3:Srinidhi_2022,C3:wang2021transpath,YANG2022102539,SimTriplet,kang2023benchmarking,C3:wang2022transformer,li2021dual,liu2023multiple}. Commonly, these methods break WSIs into smaller patches, feed them to an encoder, and extract representation features for downstream tasks. However, imbalanced category annotations within WSIs can lead to unstable performance, especially when random cropping generates image patches. While contrastive learning methods~\cite{C3:ciga2021self,li2021dual,liu2023multiple,c3:Srinidhi_2022,SimTriplet} have been applied to address this issue, it is important to note that even contrastive learning methods are not immune to imbalanced datasets \cite{selfdamage}. Consequently, several SSL methods require further improvements to seamlessly integrate into clinical practice.

In this paper, we propose a simple yet efficient self-supervised learning method called Nearby Patch Contrastive Learning (NearbyPatchCL) that treats adjacent patches as positive samples in a supervised contrastive framework, which makes the training process more robust (See Fig.~\ref{fig:positive}). To grapple with the intricate challenge of imbalanced data, we adopt the decoupled contrastive learning (DCL) loss~\cite{decoupled}. The amalgamation of these approaches not only enhances the overall effectiveness and stability of our methodology but also engenders a strong representation that holds substantial promise for clinical integration. Remarkably, our approach yields commendable performance even when furnished with a scant fraction of labeled data, demonstrating its potential utility in real-world applications within the medical domain.

For patch-level multi-class evaluation, we adopt and process WSIs from the public CAnine CuTaneous Cancer Histology (CATCH)~\cite{CATCH}, resulting in a new benchmark dataset, namely P-CATCH. Intensive experiments on the newly constructed P-CATCH dataset demonstrate the superiority of NearbyPatchCL. Our method achieves the top-1 classification accuracy of 87.56\% and significantly outperforms the supervised baseline and existing SSL methods. The proposed method also achieves compatible results when using only 1\% labeled data compared to others using  100\% labeled data. The source code will be released upon the paper's acceptance. Our contributions are summarized in the following:
\begin{itemize}
    \item We propose a novel self-supervised method, namely NearbyPatchCL, which incorporates the modified supervised contrastive loss function by leveraging nearby patches as positive samples combined with the decoupled contrastive learning loss for better representation.
    
    \item We introduce a new dataset for benchmarking WSI patch-level multi-class classification methods.
    
    \item We perform a comprehensive comparison with state-of-the-art SSL methods to demonstrate the superior performance of the proposed method. 
\end{itemize}

\section{Related works}
\subsection{Self-Supervised Learning}

SSL methods can be separated into four types based on their learning techniques. 
\textbf{Contrastive learning} algorithms (e.g., SimCLR \cite{simclr} and MoCo \cite{moco}) aim to distinguish individual training data instances from others, creating similar representations for positive pairs and distinct representations for negative pairs. However, these methods require diverse negative pairs, often mitigated by using large batches or memory banks. Yeh et al.\cite{decoupled} introduced decoupled contrastive learning loss, enhancing learning efficiency by removing the positive term from the denominator. This method can achieve competitive performance with reduced sensitivity to sub-optimal hyperparameters, without the need for large batches or extended training epochs. 
Additionally, \textbf{asymmetric networks}, such as BYOL \cite{byol} and SimSiam \cite{simsiam}, exhibit parallels with contrastive learning methods as they both learn representations of images from various augmented viewpoints. BYOL uses 2 distinct networks to create such representations, while SimSiam utilizes Siamese Networks\cite{siamese}. A distinguishing factor compared to contrastive methods is that these strategies operate independently of negative pair incorporation, enabling them to function effectively even when dealing with small batch sizes. 
On the other hand, \textbf{clustering-based methods}, including DeepCluster \cite{C3:DeepCluster} and SwAV \cite{Swav}, pursue the discovery of meaningful and compact representations by leveraging the notion of similarity and dissimilarity between data points. These methods operate under the assumption that similar data points should be closer together in the embedding space, while dissimilar points should be far apart. 
Meanwhile, \textbf{feature decorrelation} methods, exemplified by VICReg \cite{vicreg} and BarlowTwins \cite{barlowtwins}, address redundancy among different dimensions of learned features, preventing collapse or over-reliance on specific dimensions. By reducing redundancy, these methods enable more reliable and comprehensive representations, contributing to advancements in self-supervised learning techniques and achieving results on par with state-of-the-arts on several downstream tasks.

\subsection{SSL in Digital Pathology Images Analysis}

By leveraging the inherent structure and relationships within the data, SSL techniques can learn rich and meaningful representations without relying on explicit annotations. Ciga et al. \cite{C3:ciga2021self} highlighted the benefit of amalgamating diverse multi-organ datasets, including variations in staining and resolution, along with an increased number of images during the SSL process for enhanced downstream task performance. They achieved impressive results in histopathology tasks like classification, regression, and segmentation. Srinidhi et al. \cite{c3:Srinidhi_2022} introduced a domain-specific contrastive learning model tailored for histopathology, which outperforms general-purpose contrastive learning methods on tumor metastasis detection, tissue type classification, and tumor cellularity quantification. They focused on enhancing representations through a pretext task that involves predicting the order of all feasible sequences of resolution generated from the input multi-resolution patches. Wang et al.~\cite{C3:wang2021transpath} developed a SSL approach integrating self-attention to learn patch-level embeddings, called semantically-relevant contrastive learning, using convolutional neural network and a multi-scale Swin Transformer architecture as the backbone, which compares relevance between instances to mine more positive pair. The SSL method proposed by Yang et al. \cite{YANG2022102539} comprises two self-supervised stages: cross-stain prediction and contrastive learning, both grounded in domain-specific information. It can merge advantages from generative and discriminative models. In addition, SimTriplet \cite{SimTriplet} proposed by Liu et al. uses the spatial neighborhood on WSI to provide rich positive pairs (patches with the same tissue types) for triplet representation learning. It maximizes both intra-sample and inter-sample similarities via triplets from positive pairs, without using negative samples.  Also, the benchmarking created by Kang et al. \cite{kang2023benchmarking}, which includes MOCO \cite{moco}, BarlowTwin \cite{barlowtwins}, SwAV \cite{Swav}, and DINO \cite{DINO}, concluded that large-scale domain-aligned pre-training is helpful for pathology, showing its value in scenarios with limited labeled data, longer fine-tuning schedules, and when using larger and more diverse datasets for pre-training. 

Our proposed approach is related to SimTripet \cite{SimTriplet}. By leveraging a certain number of nearby images in the self-supervised training process, we aim for a robust patch-wise representation for achieving better results on patch-level multi-class datasets.



\section{Methodology}

\subsection{Overview}

Inheriting the similar structure from Supervised Contrastive Learning (SupCon)~\cite{SupCon}, our method aims to learn visual representations by multi-positive samples. Differently from previous contrastive learning methods \cite{simclr,moco} treating one augmented view as a positive sample for the other, we harness label information to add more positives for one sample. However, labels are not provided in the self-supervised setting like SupCon~\cite{SupCon}; so we adopt nearby patches as alternative positives as illustrated in Fig. \ref{fig:positive}. 

\begin{figure}[!t]
    \centering
    \includegraphics[width=0.6\textwidth]{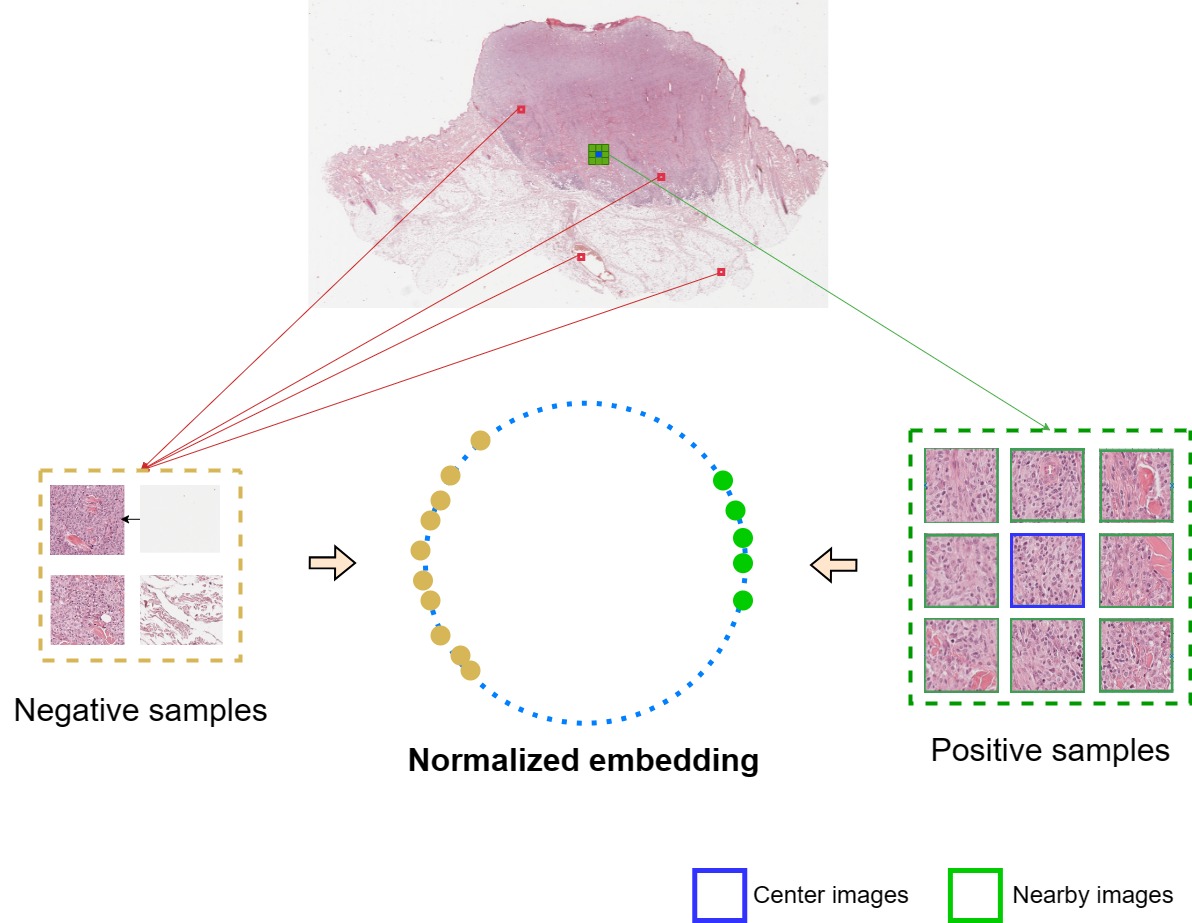}
    \caption{Overview of the proposed NearbyPatchCL. Normalized embeddings of nearby images are pulled closer to their center while pushing away other images.}
    \label{fig:overview}
\end{figure}

To this end, we propose a novel Nearby Patch Contrastive Learning (NearbyPatchCL), where we maximize the similarity between not only different views of the same patch image but also adjacent patches (See Fig. \ref{fig:overview}). Inspired from SimTriplet \cite{SimTriplet}, in our method, neighbor patches share the same tissue (or class) because they are cropped at a small scale of WSI.

\begin{figure}[!t]
    \centering
    
    \includegraphics[width=0.68\linewidth]{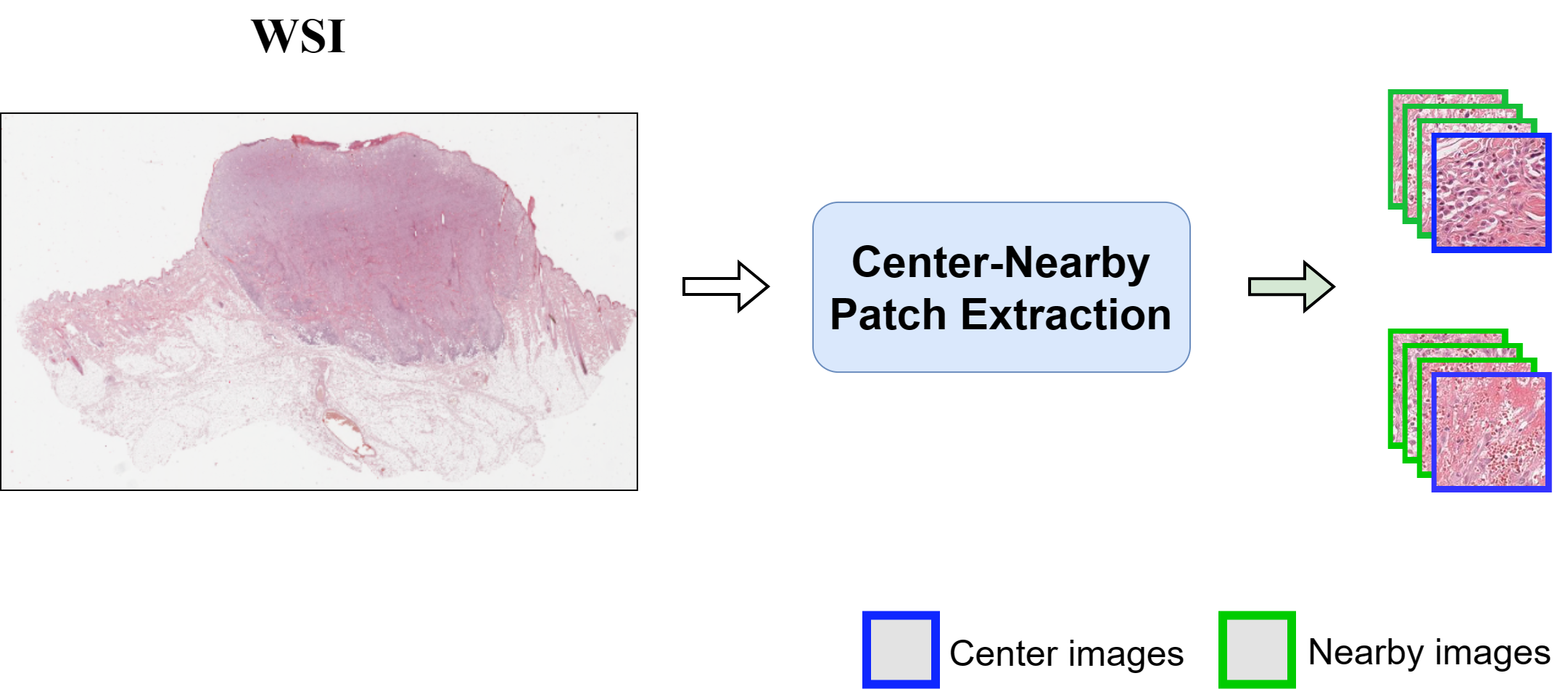}\\
    \small (a) \textbf{Extract center-nearby patches}. \\

    \vspace{10pt}
    
    \includegraphics[width=0.8\linewidth]{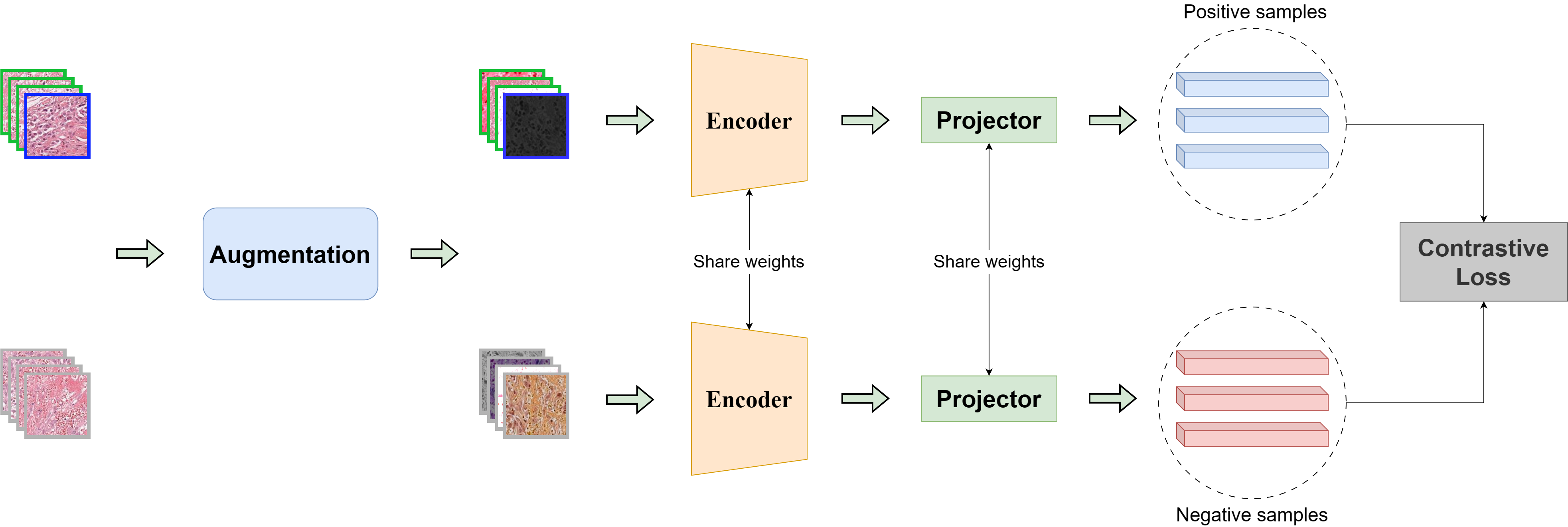} \\
    \small (b) \textbf{Encode features and minimize loss}.
    
    \label{fig:architecture}
    \caption{The architecture of NearbyPatchCL includes two main parts: (a) Extract center patches with 8 corresponding neighbors from WSIs which is a preprocessing step discussed in section \ref{sec:dataset} and (b) A contrastive loss is used to minimize the distance between learned features for positive samples (center-nearby) while simultaneously maximizing the feature distance from negative samples.}
\end{figure}







\subsection{Nearby Patch Contrastive Learning (NearbyPatchCL)}

Given a randomly sampled batch of $B$ samples (denoted as "batch"), each image is transformed by two random augmentations to get the training batch containing $2B$ samples (denoted by "multiviewed batch"). 
Besides, we denote $I \equiv \{1, \ldots, 2B\}$ as the set of indices of all samples in multiviewed batch. For a sample with index $i$, $P(i)$ and $A(i) \equiv I \backslash P(i)$ are the collection of indices of all positives and negatives related to the sample respectively, and $j(i)$ is the index of the other augmented sample originating from the same source sample in the batch.





\begin{algorithm}[!t]
    \caption{Pseudocode for NearbyPatchCL algorithm.}
    \label{alg:pseudo}
    \begin{algorithmic}[1]
        \Statex {
        \Input 
            \Desc{X}{Sampled minibatch}
            \Desc{C}{Number of center samples in $X$}
            \Desc{I}{Index set of samples in the multiviewed batch}
            \Desc{N}{Number of nearby samples per center}
            \Desc{$\mathcal{N}_n(x_i)$}{Function return n-th nearby of sample $x_i$}
            \Desc{$\mathcal{T}$}{Distribution of image transformation}
            \Desc{$f$, $g$}{Encoder network and Projection head}
            \Desc{$P(i)$}{Set of indices of positive samples for sample with index $i$.}
            \Desc{$A(i)$}{Set of indices of negative samples for sample with index $i$.}
        \EndInput
        }
        
        \For{sampled minibatch $X = \{x_c\}_{c=1}^C$}
            \State $Y = \{1, \ldots, C\}$
            \Comment{labels}
            \For{$n \in \{1, \ldots, N\}$}\Comment{Retrieve N nearby samples}
                \For{$c \in \{1 \ldots C\}$}
                    \State $X \gets X + \mathcal{N}_n(x_c)$
                    \State $Y \gets Y + c$
                \EndFor
            \EndFor

        
            \State $t \sim \mathcal{T}$, $t^{\prime} \sim \mathcal{T}$ \Comment{two augmentations}

            \For{$k \in \{1, \ldots, C(N+1)\}$}
                \State $\tilde{x}_{k} = t(x_k)$
                \Comment{first augmentation}
                \State $h_k$ = $f(\tilde{x}_k)$
                \State $z_k = g(h_k)$

                
                \State $\tilde{x}_{k+C(N+1)} = t^{\prime}(x_k)$
                \Comment{second augmentation}
                \State $h_{k+C(N+1)}$ = $f(\tilde{x}_{k+C(N+1)})$
                \State $z_{k+C(N+1)} = g(h_{k+C(N+1)})$
            \EndFor


            \For{$i \in \{1, \ldots, 2C(N+1)\}$ and $j \in \{1, \ldots, 2C(N+1)\}$}
                \State $s_{i,j} = z_i^T z_j / \| z_i \| \| z_j \|$
                \Comment{pairwise similarity}
            \EndFor

            \For{$i \in I$}
                \State $l(i) = \frac{-1}{|P(i)|} \sum\limits_{p \in P(i)} \log \frac{\exp \left( z_i \cdot z_p \backslash \tau \right)}{\sum\limits_{a \in A(i)} \exp \left( z_i \cdot z_a \backslash \tau \right)}$ 
            \EndFor

            \State $\mathcal{L} = \frac{1}{2C(N+1)} \sum_{i=1}^{2C(N+1)} l(i) $

            \State update networks f and g to minimize $\mathcal{L}$
        
        \EndFor

        \Output \hspace{2em} Encoder $f$
        \EndOutput
    \end{algorithmic}
\end{algorithm}

\vspace{-10pt}




Each batch contains $C$ center patches with $N$ corresponding nearby patches ($0 \leq N \leq 8$), so there are $B = C \times (N + 1)$ samples, $\{x_k\}_{k=1 \ldots B}$, in total. A group of patches including a center and corresponding neighbors belongs to one class; therefore, there are $C$ classes in the batch. By applying two transformations, the multiviewed batch contains $2C(N+1)$ samples, $\{\widetilde{x}_l\}_{l=1 \ldots 2B}$. For an arbitrary sample with index $i$, $P(i) \equiv \{j(i)\} \cup \mathcal{N}(i)$ where $\mathcal{N}(i)$ is the set that consists of $2N$ indices of positives (nearby) for the sample and $2(N+1)(C-1)$ remaining instances as negatives. Our naive loss function is as follows:
\begin{equation}
    \label{eq:main_loss}
    \mathcal{L}= \sum_{i \in I}\frac{-1}{|{P}{(i)}|}  \sum_{p \in P(i)} \log \frac{\exp \left(z_i \cdot z_p / \tau \right)}{\exp \left(z_i \cdot z_p / \tau \right) +  \sum\limits_{a \in A(i)} \exp \left(z_i \cdot z_a / \tau \right)},
\end{equation}
where ${z}_{i} \equiv g\left(f\left(\tilde{x}_i\right)\right)$ is the output feature of sample $\tilde{x}_i$, the $\cdot$ symbol denotes the inner product, and $\tau$ is a scalar temperature parameter. 

Inspired by Yeh et al. \cite{decoupled}, to address the imbalance due to cropping WSIs, we hypothesize that robustness representations can be obtained by removing negative samples in a batch to eliminate the negative pair contrast effect. It means that we do not need a large batch size during the training process to enhance learning efficiency, resulting in stable performance. This hypothesis is explored further in Section \ref{ablation:decoupled}. Our naive loss function (Eq. \ref{eq:main_loss}) becomes:

\begin{equation}
    \label{eq:final_loss}
    \mathcal{L}= \sum_{i \in I}\frac{-1}{|{P}{(i)}|}  \sum_{p \in P(i)} \log \frac{\exp \left(z_i \cdot z_p / \tau \right)}{\sum\limits_{a \in A(i)} \exp \left(z_i \cdot z_a / \tau \right)}. 
\end{equation}

Algorithm \ref{alg:pseudo} summarizes our proposed method.

\subsection{Implementation Details}

\subsubsection{Architecture.}

Figure \ref{fig:architecture} shows the overall architecture of our NearbyPatchCL, adopted from SupCon's architecture~\cite{SupCon}. ResNet-50 backbone is used. The representation from ResNet-50's last fully connected layer remains 2048-dimensional, and it's then reduced to 128 dimensions using a multi-layer perceptron (MLP).

\subsubsection{Image augmentations.}

We follow the augmentation described in SupCon~\cite{SupCon}. Additionally, horizontal flips are randomly applied for rotation invariance in skin tissue images. Images are resized to $128 \times 128$ for computational efficiency.  

\subsubsection{Optimization.}

We use the SGD optimizer, with a base learning rate of $0.2$, momentum of $0.9$, and weight decay of $0.0001$. To optimize resources, we use Mixed Precision Training (MPT) \cite{MP} with a learning rate scheduler. The new learning rate is calculated as $lr \times BatchSize \times (N+1) / 256$, where  $BatchSize = 512/(N+1)$ is the number of center images and $N$ is the number of nearby patches per center. The scheduler involves 10 warm-up epochs followed by cosine decay. This SSL setup takes about 73 hours to train the encoder for 400 epochs using unlabeled data from our newly constructed P-CATCH dataset.



\section{Proposed P-CATCH Dataset}
\label{sec:dataset}

\subsection{Original CATCH Images}

The CATCH dataset \cite{CATCH} consists of 350 WSIs, stored in the pyramidal Aperio file format, having direct access to three resolution levels (0.25$\frac{\mu m}{px}$, 1$\frac{\mu m}{px}$, 4$\frac{\mu m}{px}$). 
Pathologists used Slide Runner software to create a database with 12,424 area annotations in total, covering six non-neoplastic tissues (epidermis, dermis, subcutis, bone, cartilage, inflammation/necrosis) and seven tumor classes. Notably, there is a significant imbalance in the distribution of the six non-neoplastic tissue classes within the database.


\begin{table}[t!]
    \caption{The number of patch images in unlabeled sets, each has 247 WSIs. The number after the dataset name denotes the number of nearby images.}
    \centering
    \resizebox{1\linewidth}{!}{
    \begin{tabular}{|l|r|r|r|r|}
        \hline
        \multicolumn{1}{|c|}{\multirow{2}{*}{\textbf{Unlabeled Set}}} & \multicolumn{2}{|c|}{\textbf{Per WSI}} & \multicolumn{2}{|c|}{\textbf{Total}} \\ \cline{2-5}
        \multicolumn{1}{|c|}{} & \multicolumn{1}{c|}{\textbf{Center Images}} & \multicolumn{1}{c|}{\textbf{Nearby Images}} & \multicolumn{1}{c|}{\textbf{Center Images}} & \multicolumn{1}{c|}{\textbf{Nearby Images}} \\ \hline
        P-CATCH-0 & 1,000 & 0 & 247,000 & 0 \\ \hline
        P-CATCH-1 & 500 & 500 & 123,500 & 123,500 \\ \hline
        P-CATCH-2 & 333 & 666 & 82,251 & 164,502 \\ \hline
        P-CATCH-4 & 200 & 800 & 49,400  & 197,600 \\ \hline
        P-CATCH-8 & 111 & 889 & 27,417 & 246,753 \\ \hline
    \end{tabular}
    }
    \label{datasets}
\end{table}

\begin{table}[t!]
    \caption{The number of patch images per category in our P-CATCH dataset.}
    \centering
    \begin{tabular}{|c|r|r|c|}
        \hline
        \textbf{Category} & \textbf{Training Set} & \textbf{Test Set} & \textbf{Unlabeled Set} \\
        \hline
        Dermis & 22,020 & 81,143 & - \\
        \hline
        Epidermis & 9,471 & 16,001 & - \\
        \hline
        Inflamm/Necrosis & 19,488 & 24,612 & - \\
        \hline
        Subcutis & 16,566 & 53,426 & - \\
        \hline
        Tumor & 22,341 & 96,188 & - \\
        \hline
        Background & 1,917 & 4,533 & - \\
        \hline
        Total & 91,803 & 175,903 & 2,223,000 \\
        \hline
    \end{tabular}
    \label{category}
\end{table}

\subsection{Proposed P-CATCH Dataset}

\subsubsection{Unlabeled sets.} A total of 70\% of WSIs contained within the CATCH database, specifically corresponding to 247 WSIs, is used as unlabeled data for training SSL methods. To ensure a balanced representation of the diverse tumor subtypes present, we adopt an equitable distribution strategy, resulting in approximately 35 WSIs per subtype. From each WSI at the 0.25$\frac{\mu m}{px}$ resolution level, we randomly extract image patches with size of $512 \times 512$ pixels, encompassing center patches along with nearby patches. We create different subsets, denoted by P-CATCH-$N$, where $N$ indicates the number of nearby samples of a center image. Table \ref{datasets} shows the statistics of created five subsets, corresponding to 0, 1, 2, 4, and 8 nearby samples. We remark that we try to ensure an equal number of patch images for P-CATCH-$N$ subsets. 



\subsubsection{Annotated sets.} We partitioned the dataset, with the remaining 97 WSIs reserved for the test set, while 37 WSIs from the unlabeled set were allocated for training purposes. Specifically, we conduct random sampling to extract non-overlapping image patches, each size at $512 \times 512$ pixels, from the WSIs at the 0.25$\frac{\mu m}{px}$ resolution level. This process yields approximately 92,000 images distributed across six categories for the training set, and around 176,000 images for the test set, spanning the same set of categories. The distribution of images across these categories is detailed in Table \ref{category}. This meticulous dataset partitioning ensures a rigorous and comprehensive evaluation of our methodology.



\section{Experiments}

\subsection{Linear Evaluation Protocol}

We adhere to the well-established linear evaluation protocol~\cite{simclr,barlowtwins} which entails training a linear classifier atop a frozen base network (i.e., ResNet-50), and test accuracy is used as a proxy for representation quality. A single transformation is applied, involving resizing to $256 \times 256$, subsequent center-cropping to revert to the original resolution of $224 \times 224$, followed by normalization. Methods are trained for 15 epochs, utilizing SGD optimizer with a learning rate of $0.2$, momentum of $0.9$, and weight decay of $0$. The batch size is configured at 32 for training on 1\% labeled data, while a batch size of 512 is employed for cases involving 10\%, 20\%, and 100\% labeled data. To ensure a robust and comprehensive assessment, we employ 5-fold cross-validation on the training set, resulting in 5 trained models. Notably, MPT is excluded during the linear evaluation. During the evaluation of the test set, the results obtained from these 5 models are averaged to provide a more reliable estimate of the model's performance, thereby effectively mitigating issues such as overfitting and yielding a more robust evaluation outcome.





\subsection{Comparison with State-of-the-arts}

We compared our NearbyPatchCL, using $N=4$ nearby samples, with state-of-the-art SSL methods, including SimCLR \cite{simclr}, SimSiam \cite{simsiam}, BarlowTwins \cite{barlowtwins}, SimTriplet \cite{SimTriplet}, and BYOL \cite{byol}. For fair evaluation, all methods employed the ResNet-50 backbone pre-trained on the ImageNet dataset. SimTriplet \cite{SimTriplet} used P-CATCH-1 for self-supervised training phase. Meanwhile, P-CATCH-0 subset was used for training other SSL methods. Training parameters follow their original work. 


\begin{table}[!t]
\centering
  \caption{Classification performance, using different amounts of training data. With different numbers of annotated data, we ensure different classes contribute similar numbers of images to address the issue that the annotation is highly imbalanced.}
  \label{tab:per_accuracy}
  
  \begin{tabular}{|l|r|r|r|r|r|r|r|r|}
    \hline
    \textbf{Method} & \multicolumn{4}{|c|}{\textbf{F1 Score}} & \multicolumn{4}{|c|}{\textbf{Balanced Accuracy}} \\
    \cline{2-9}
    & \textbf{1\%} & \textbf{10\%} & \textbf{20\%} & \textbf{100\%} & \textbf{1\%} & \textbf{10\%} & \textbf{20\%} & \textbf{100\%} \\
    
    \hline 

    Supervised baseline  & 66.95 & 75.92  & 78.23  & 81.04 & 74.79 & 81.94 & 83.58 & 84.24 \\

    SimCLR \cite{simclr} & 70.09 & 72.42 & 72.43 & 72.32 & 77.55 & 79.45 & 79.40 & 79.26 \\
    
    SimSiam \cite{simsiam} & 54.24 & 57.52 & 57.60 & 58.43 &  58.59 & 60.14 & 60.69 & 61.31 \\

    SimTriplet \cite{SimTriplet} & 55.18 & 57.11 & 56.20  & 58.67 & 60.75 & 62.65 & 62.78  & 63.57 \\

    BYOL \cite{byol} & 65.62 & 77.35 & 79.24 & 83.01 & 74.63 & 83.41 & 84.51 & 85.48 \\
    
    BarlowTwins \cite{barlowtwins} & \underline{76.43} & \underline{80.88} & \underline{81.08} &\underline{82.55} & \underline{82.73} & \underline{85.98} &\underline{86.19} & \underline{86.36}\\

    NearbyPatchCL (N=4) & \textbf {81.85} & \textbf {83.73} & \textbf {84.41} & \textbf {85.72} & \textbf {85.63} & \textbf {87.19} &\textbf {87.56} & \textbf{87.14} \\
    \hline
  \end{tabular}
\end{table}

\begin{figure*}[t!]
    \centering
    
    \begin{tabular}{@{} c @{\hspace{0.0cm}} c @{\hspace{0.0cm}} c @{}}
        \includegraphics[width=0.25\textwidth]{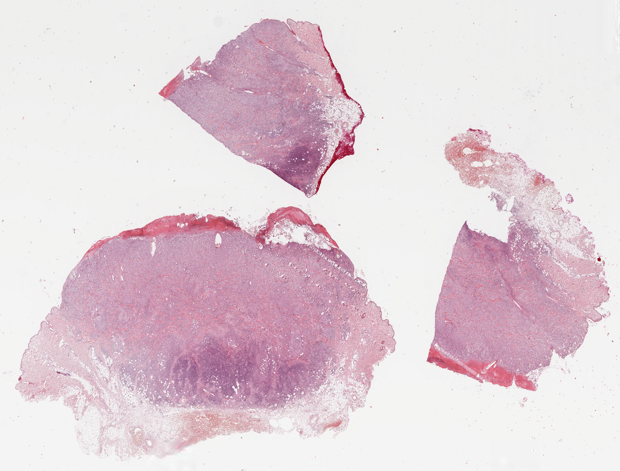} & \includegraphics[width=0.25\textwidth]{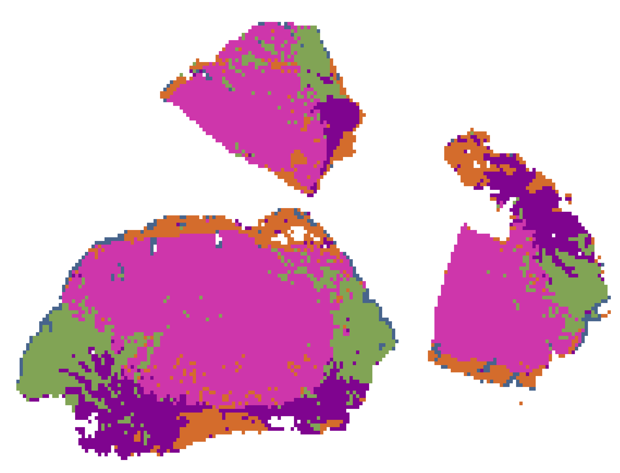} &
        \includegraphics[width=0.25\textwidth]{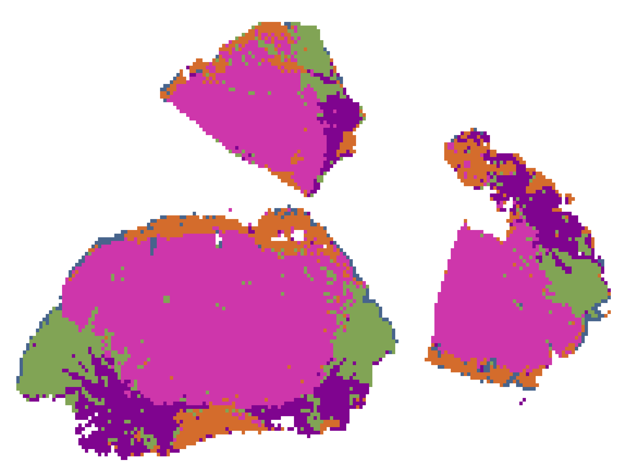}\\
        [\abovecaptionskip] \small (a) Origin image & (b) NearPatchCL(100\%) & (c) NearPatchCL(1\%)
    \end{tabular}

    \begin{tabular}{@{} c @{\hspace{0.0cm}} c @{\hspace{0.0cm}} c @{}}
        \includegraphics[width=0.25\textwidth]{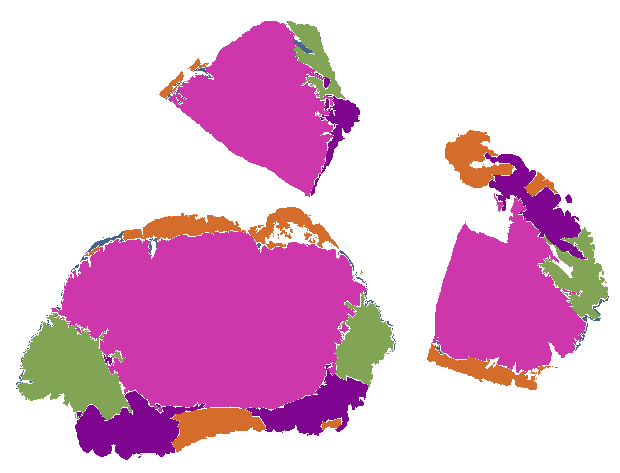} & \includegraphics[width=0.25\textwidth]{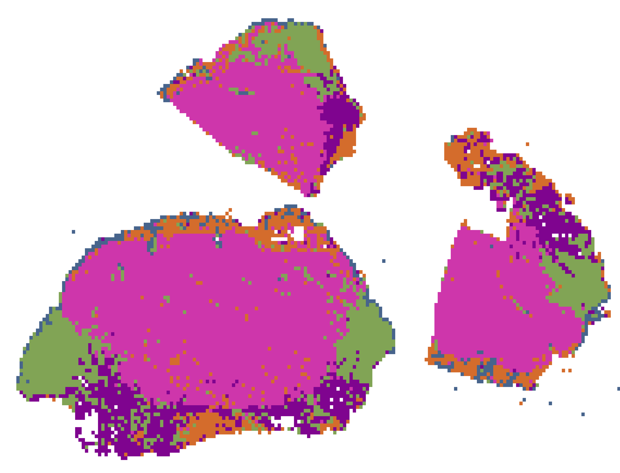} &
        \includegraphics[width=0.25\textwidth]{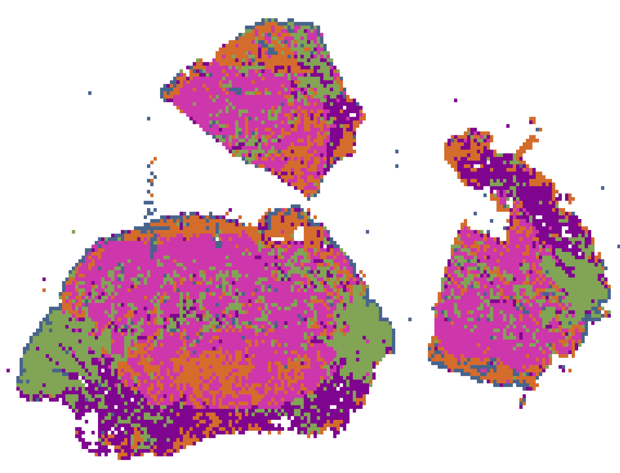}\\
        [\abovecaptionskip] \small (c) Manual annotation & (d) Supervised(100\%) & (e) Supervised(1\%)
    \end{tabular}
    \begin{tikzpicture}[remember picture, overlay]
        \node [anchor=west] at (-11.1, -2.5) {\textbf{Label:}};
        
        \draw [Tumor, line width=3mm] (-9.8, -2.5) -- (-9.3,-2.5);
        \node [anchor=west] at (-9.4, -2.5) {Tumor};
        
        \draw [Dermis, line width=3mm] (-8, -2.5) -- (-7.5, -2.5);
        \node [anchor=west] at (-7.6, -2.5) {Dermis};
        
        \draw [Subcutis, line width=3mm] (-6.1, -2.5) -- (-5.6, -2.5);
        \node [anchor=west] at (-5.7, -2.5) {Subcutis};
        
        \draw [Epidermis, line width=3mm] (-4, -2.5) -- (-3.5, -2.5);
        \node [anchor=west] at (-3.6, -2.5) {Epidermis};
        
        \draw [Inflamm-Necrosis, line width=3mm] (-1.8, -2.5) -- (-1.3, -2.5);
        \node [anchor=west] at (-1.3, -2.5) {Inflamm-Necrosis};
    \end{tikzpicture}
    
\vspace*{+10mm}
    \caption{Visualization of classification results on a sample WSI from the test set. Our proposed method shows significantly better results than that of the supervised baseline, even with only $1\%$ of the training data.}
    \label{fig:demo}
\end{figure*}

Comparison results in Table \ref{tab:per_accuracy} demonstrate that our NearbyPatchCL(N=4) significantly outperforms the supervised baseline (i.e., frozen ImageNet pre-trained ResNet-50 with a classifier trained on the training set) and other SSL methods across all data proportions. Furthermore, even with 1\% labeled data used for training, NearbyPatchCL still has competitive results compared to BYOL and Barlowtwins using 100\% of training data. 
Leveraging nearby batches as positive samples, our approach achieves better results than state-of-the-art.
Additionally, we visualize results on a whole WSI in Fig. \ref{fig:demo}. The result also indicates that our approach can effectively leverage unlabeled data for improving 
classification tasks, making them a valuable tool in practical scenarios where annotating WSIs is costly.

\subsection{Ablation Study}

\subsubsection{Number of nearby samples.}
\label{ablation:nearby}

We investigate the influence of the number of nearby samples ($N$) for each center image on the performance of the proposed method. It is notable that with N=0, NearbyPatchCL becomes SimCLR. The result in Table \ref{tab:per_accuracy_decouple}, with DCL loss, shows that the performance of NearbyPatchCL increases gradually from $N=0$ to $N=4$ and achieves the best overall performance at $N=4$ in using 10\%, 20\%, and 100\% labeled images in the training set. The result in NearbyPatchCL(N=8) still has a competitive result with NearbyPatchCL(N=4) and has the second-highest overall performance. It is notable that annotations made by pathologists may not have an accuracy of 100\%. Hence, training on a large amount of data can make the model learn some false cases, as we can see the balanced accuracy drop when moving from $20\%$ to $100\%$ in $N=1,4,8$. On the other hand, the performance without DCL loss increasing gradually from $N=0$ to $N=8$ also shows the superiority of utilizing nearby patches in the SSL process.

\subsubsection{Effect of DCL Loss.}
\label{ablation:decoupled}

\begin{table}[!t]
    \caption{Ablation study with different numbers of nearby samples (N) for each image center and using DCL loss.}
    \centering
    \begin{tabular}{|l|c|r|r|r|r|r|r|r|r|}
        \hline
        \multicolumn{2}{|c|}{\multirow{2}{*}{\textbf{Method}}} & \multicolumn{4}{|c|}{\textbf{F1 Score}} & \multicolumn{4}{|c|}{\textbf{Balanced Accuracy}} \\
        \cline{3-10}
        
         \multicolumn{2}{|c|}{} & \textbf{1\%} & \textbf{10\%} & \textbf{20\%} & \textbf{100\%} & \textbf{1\%} & \textbf{10\%} & \textbf{20\%} & \textbf{100\%} \\
         \hline
         
        NearbyPatchCL (N=0) & \multirow{5}{*}{DCL}& 73.83 & 78.36 & 79.14     & 80.69 & 80.01 & 83.87 & 84.35 & 84.44 \\

        NearbyPatchCL (N=1) & & 77.97 & 81.84 & 82.77 & 84.40 & 83.42 & 86.54 & 87.00 & 86.94 \\

        NearbyPatchCL (N=2) & & 80.27 & 82.87 & 83.43 & 85.07 & \underline{85.63} & 87.16 & 87.38 & \textbf{87.47} \\

         NearbyPatchCL (N=4) & & \underline{81.85} & \textbf{83.73} & \textbf{84.41} & \textbf{85.72} & \underline{85.63} & 87.19 & \textbf{87.56} & \underline{87.14} \\
         
         NearbyPatchCL (N=8) &  & 79.66 & 83.14 & 83.97 & \underline{85.45} & 84.60 & \textbf{87.29} & \underline{87.53} & 87.09 \\

         \hline

         NearbyPatchCL (N=0) & \multirow{5}{*}{w/o DCL}& 70.09 & 72.42 & 72.43 & 72.32 & 77.55 & 79.45 & 79.40 & 79.26 \\

         NearbyPatchCL (N=1) & & 77.52 & 82.58 & 83.31 & 84.68 & 83.93 & 86.92 & 87.12 & 86.57 \\

         NearbyPatchCL (N=2) & & 77.69 & 82.41 & 83.25 & 84.82 & 82.85 & 86.42 & 86.86 & 86.86 \\

         NearbyPatchCL (N=4) & & 78.62 & 82.68 & 83.62 & 84.94 & 83.77 & 86.90 & 87.40 & 87.08 \\
         
         NearbyPatchCL (N=8) &  & \textbf{82.17} & \underline{83.58} & \underline{84.00} & 84.41 & \textbf{85.86} & \underline{87.22} & 87.47 & 86.67 \\

         \hline
    \end{tabular}
    \label{tab:per_accuracy_decouple}
\end{table}


The result shown in Table \ref{tab:per_accuracy_decouple} indicates that with more nearby samples for each center image, as we can see in $ N=\{2,4\}$, employing the DCL loss can improve the performance with all percentages of labeled data. 

Although our method, with DCL loss, achieves the best overall performance at $N=4$, NearbyPatchCL (w/o DCL) still achieves the best results in 1\% and 10\% of training data. With $N=1$ or $8$, NearbyPatchCL with DCL loss does not have any significant improvement in performance, even worse at some percentages of labeled data as in $N=1$. Our hypothesis is that NearbyPatchCL gains more benefit from leveraging more nearby patches with DCL loss. However, with $N=8$, there are only 111 center images that are randomly cropped from each WSI, which are nearly half of that of $N=4$, this can make the training data more unbalanced due to the differences in area and quantity of each category annotation, leading to a decrease in overall performance from $N=4$ to $N=8$. 

\section{Conclusion and Future works}

In this paper, we have conducted a new benchmark of patch-level multi-class WSI classification using SSL methods. To tackle the scarcity of labeled data and imbalanced datasets issue in digital pathology images, our work has shown that by leveraging nearby patches as positive samples in the SSL phase, the proposed method can have a more robust representation and perform better on downstream tasks. Furthermore, we have shown that using DCL loss can benefit contrastive methods while training on an imbalanced dataset.

In future work, we aim to extend our approach to other medical imaging domains and explore its application in other downstream tasks. We also plan to further investigate methods to enhance the interpretability of learned representations and incorporate domain-specific knowledge to improve the performance of the model in real-world clinical settings.

\section*{Acknowledgement}
This research was funded by Vingroup and supported by Vingroup Innovation Foundation (VINIF) under project code VINIF.2019.DA19. This project was also supported by the Faculty of Information Technology, University of Science, Vietnam National University - Ho Chi Minh City.

%
%
%


\bibliographystyle{splncs04}
\bibliography{ref}
%




\end{document}